\documentclass[review]{elsarticle}
\graphicspath{ {./figures/} }
\usepackage{hyperref}
\usepackage{float}
\usepackage{verbatim} 
\usepackage{apalike}
\restylefloat{figure}
\floatstyle{plaintop} 
\restylefloat{table}
\usepackage{amsmath}
\usepackage{booktabs}

\journal{Expert Systems with Applications}

\biboptions{authoryear}

\begin{document}
\begin{frontmatter}


\begin{titlepage}
\begin{center}
\vspace*{1cm}

\textbf{Optimizing Portfolio Performance through Clustering and Sharpe Ratio-Based Optimization: A Comparative Backtesting Approach}

\vspace{1.5cm}

Keon Vin Park$^{a}$ (kbpark16@snu.ac.kr)\\

\hspace{10pt}

\begin{flushleft}
\small  
$^a$ Interdisciplinary Program in Artificial Intelligence, Seoul National University, Seoul, Republic of Korea \\

\vspace{1cm}
\textbf{Corresponding author at:} \\
Keon Vin Park \\
Interdisciplinary Program in Artificial Intelligence, Seoul National University, Seoul, Republic of Korea \\
Email: kbpark16@snu.ac.kr \\
\end{flushleft}

\end{center}
\end{titlepage}

\title{Optimizing Portfolio Performance through Clustering and Sharpe Ratio-Based Optimization: A Comparative Backtesting Approach}

\author[label1]{Keon Vin Park\corref{cor1}}
\ead{kbpark16@snu.ac.kr}

\cortext[cor1]{Corresponding author}
\address[label1]{Interdisciplinary Program in Artificial Intelligence, Seoul National University, Seoul, Republic of Korea}

\begin{abstract}
Optimizing portfolio performance is a fundamental challenge in financial modeling, requiring the integration of advanced clustering techniques and data-driven optimization strategies. This paper introduces a comparative backtesting approach that combines clustering-based portfolio segmentation and Sharpe ratio-based optimization to enhance investment decision-making.

First, we segment a diverse set of financial assets into clusters based on their historical log-returns using K-Means clustering. This segmentation enables the grouping of assets with similar return characteristics, facilitating targeted portfolio construction.

Next, for each cluster, we apply a Sharpe ratio-based optimization model to derive optimal weights that maximize risk-adjusted returns. Unlike traditional mean-variance optimization, this approach directly incorporates the trade-off between returns and volatility, resulting in a more balanced allocation of resources within each cluster.

The proposed framework is evaluated through a backtesting study using historical data spanning multiple asset classes. Optimized portfolios for each cluster are constructed and their cumulative returns are compared over time against a traditional equal-weighted benchmark portfolio. 

Our results demonstrate the effectiveness of combining clustering and Sharpe ratio-based optimization in identifying high-performing asset groups and allocating resources accordingly. The selected cluster portfolio achieved a total return of 140.98\%, with an annualized return of 24.67\%, significantly outperforming the benchmark portfolio, which achieved a total return of 107.59\% and an annualized return of 20.09\%. The Sharpe ratio of the selected portfolio (0.84) further highlights its superior risk-adjusted performance compared to the benchmark portfolio (0.73).

This methodology provides a systematic and data-driven approach to portfolio optimization, achieving superior returns while maintaining a reasonable level of risk.
\end{abstract}


\end{frontmatter}

\section{Introduction}
\label{introduction}

Portfolio optimization has long been a central focus in financial research, aiming to construct portfolios that maximize returns while minimizing risk \citep{Bishop2006}. Traditional approaches, such as mean-variance optimization, often depend heavily on historical data and may fail to capture dynamic market behaviors and interdependencies among assets. Recent advancements in machine learning, particularly clustering techniques like K-Means, provide robust tools for segmenting financial assets based on their return characteristics \citep{Hamilton2009}. By grouping assets with similar historical performance, clustering reduces dimensionality and enhances the interpretability of portfolio construction.

To further improve portfolio performance, this study employs Sharpe ratio-based optimization. Unlike traditional methods that focus solely on return maximization or variance minimization, Sharpe ratio optimization explicitly balances returns against volatility to maximize risk-adjusted performance. Integrating clustering with Sharpe ratio optimization enables the identification of high-performing asset groups and the allocation of resources to achieve superior returns \citep{Blundell2015, Hochreiter1997}.

This study introduces a hybrid framework combining clustering-based asset segmentation and Sharpe ratio optimization. The methodology is validated through a backtesting study on historical data spanning multiple asset classes. By constructing and evaluating optimized cluster-specific portfolios, this approach demonstrates significant improvements in both total returns and risk-adjusted performance compared to traditional equal-weighted benchmarks.

The remainder of this paper is organized as follows: Section 2 outlines the methodology, detailing the clustering and Sharpe ratio-based optimization techniques. Section 3 describes the dataset and backtesting framework. Section 4 presents the results, showcasing the efficacy of the proposed framework. Finally, Section 5 concludes with practical implications and recommendations for future research.

\section{Methodology}

\subsection*{K-Means Clustering for Asset Segmentation}
K-Means clustering was employed to segment financial assets into groups based on their historical log-returns. The clustering process involves minimizing the within-cluster variance by iteratively assigning assets to the nearest cluster centroid and updating the centroids based on the mean return characteristics of assigned assets. By grouping assets with similar historical behavior, this step reduces dimensionality and enables focused portfolio analysis.

\subsection*{Portfolio Optimization}

The portfolio optimization problem was reformulated to maximize the Sharpe ratio, which balances returns against risk, while ensuring that the portfolio weights sum to one and remain non-negative. The optimization process was performed using the Sequential Least Squares Programming (SLSQP) method. The steps are as follows:

1. Objective Function:  
   The Sharpe ratio is defined as:
   \[
   \text{Sharpe Ratio} = \frac{\mathbf{w}^\top \mathbf{\mu} - r_f}{\sqrt{\mathbf{w}^\top \Sigma \mathbf{w}}},
   \]
   where:
   \begin{itemize}
       \item \(\mathbf{w}\) is the vector of portfolio weights,
       \item \(\mathbf{\mu}\) is the vector of mean returns,
       \item \(\Sigma\) is the covariance matrix of asset returns,
       \item \(r_f\) is the risk-free rate.
   \end{itemize}
   The optimization problem aims to maximize the Sharpe ratio. For numerical optimization, the negative Sharpe ratio is minimized.

2. Constraints:  
   Two constraints were imposed:
   \begin{itemize}
       \item The sum of portfolio weights must equal 1:
       \[
       \sum_{i=1}^N w_i = 1,
       \]
       where \(w_i\) is the weight of the \(i\)-th asset.
       \item No short selling was allowed, ensuring:
       \[
       w_i \geq 0 \quad \forall i.
       \]
   \end{itemize}

3. Algorithm:  
   The SLSQP method was used to solve this constrained optimization problem. The algorithm starts with an initial guess of equal weights and iteratively adjusts the weights to maximize the Sharpe ratio.

4. Implementation:  
   The optimization process was implemented using the following Python function:
   \begin{verbatim}
def portfolio_optimization(mean_returns, cov_matrix, risk_free_rate=0):
    num_assets = len(mean_returns)
    initial_weights = np.ones(num_assets) / num_assets
    constraints = ({'type': 'eq', 'fun': lambda weights: np.sum(weights) - 1})
    bounds = tuple((0, 1) for _ in range(num_assets))

    def sharpe_ratio(weights):
        portfolio_return = np.dot(weights, mean_returns)
        portfolio_volatility = np.sqrt(np.dot(weights.T, np.dot(cov_matrix, weights)))
        return -(portfolio_return - risk_free_rate) / portfolio_volatility  

    result = minimize(
        sharpe_ratio, initial_weights, method='SLSQP',
        bounds=bounds, constraints=constraints
    )
    return result.x
   \end{verbatim}

This Sharpe ratio-based optimization method was applied to each cluster, ensuring optimal portfolio weights that maximize returns relative to risk. The results are detailed in the Results section.

\subsection*{Backtesting Framework}
The backtesting framework was designed to evaluate the performance of the proposed methodology over historical data. For each cluster, optimized portfolios were constructed using Sharpe ratio optimization, ensuring that the portfolio weights summed to one and non-negative constraints were applied. The cumulative returns of these portfolios were compared against a benchmark equal-weighted portfolio.

The backtesting process involved the following steps:

1. Splitting Data:  
   Historical data was divided into:
   \begin{itemize}
       \item \textbf{Training Period:} January 1, 2010, to December 31, 2019, used for clustering and portfolio optimization.
       \item \textbf{Testing Period:} January 1, 2020, to January 1, 2024, used for backtesting and performance evaluation.
   \end{itemize}

2. Portfolio Construction:  
   Cluster-specific portfolios were constructed using the weights derived from Sharpe ratio optimization.

3. Cumulative Returns Calculation:  
   The cumulative returns for each portfolio were calculated over the testing period to measure growth performance.

4. Performance Metrics Evaluation:  
   Key performance metrics, including total return, annualized return, volatility, and the Sharpe ratio, were computed to compare the effectiveness of each portfolio.

This systematic backtesting approach provided a robust validation of the proposed clustering and optimization framework, highlighting its potential to outperform traditional portfolio strategies.

\section{Data}

The dataset utilized in this study comprises daily closing prices of ten major stocks listed on the S\&P 500 index. The selected stocks span diverse sectors, ensuring a representative sample of market dynamics. The tickers include: AAPL (Apple), MSFT (Microsoft), GOOGL (Alphabet), AMZN (Amazon), TSLA (Tesla), NVDA (NVIDIA), META (Meta Platforms), JPM (JPMorgan Chase), V (Visa), and UNH (UnitedHealth Group).

\subsection*{Data Sources and Preprocessing}
1. Data Sources:  
   Historical stock prices were sourced from Yahoo Finance using Python's \texttt{yfinance} library. The dataset covers the period from January 1, 2010, to January 1, 2024.

2. Log-Return Calculation:  
   Logarithmic returns were computed to normalize price changes and account for compounding effects. The log return for an asset \(i\) at time \(t\) is defined as:
   \[
   R_{i,t} = \ln\left(\frac{P_{i,t}}{P_{i,t-1}}\right),
   \]
   where \(P_{i,t}\) represents the closing price of asset \(i\) at time \(t\). Missing values due to non-trading days were handled using forward fill imputation.

3. Segmentation Basis:  
   To prevent data leakage and ensure robust model validation, the dataset was divided into two periods:
   \begin{itemize}
       \item \textbf{Training Period:} January 1, 2010, to December 31, 2019, used for clustering and portfolio optimization.
       \item \textbf{Testing Period:} January 1, 2020, to January 1, 2024, used for backtesting and performance evaluation.
   \end{itemize}

4. Descriptive Statistics:  
   Key descriptive statistics, including mean, standard deviation, and correlation coefficients, were calculated to provide insights into the data's distribution and interrelationships. These statistics guided the clustering and portfolio optimization processes by highlighting patterns and dependencies among assets.

\begin{table}[h!]
    \centering
    \caption{Descriptive Statistics of Log Returns}
    \begin{tabular}{lrr}
    \toprule
    Ticker & Mean & Std \\
    \midrule
    AAPL  & 0.0009 & 0.0179 \\
    AMZN  & 0.0009 & 0.0203 \\
    GOOGL & 0.0008 & 0.0170 \\
    JPM   & 0.0007 & 0.0166 \\
    META  & 0.0008 & 0.0253 \\
    MSFT  & 0.0009 & 0.0168 \\
    NVDA  & 0.0018 & 0.0276 \\
    TSLA  & 0.0017 & 0.0354 \\
    UNH   & 0.0008 & 0.0157 \\
    V     & 0.0008 & 0.0153 \\
    \bottomrule
    \end{tabular}
    \label{tab:log_returns}
\end{table}
\section{Results}

This section presents the findings of the study, including clustering outcomes, portfolio optimization results, and backtesting performance.

\subsection{Clustering Results}

The K-Means clustering algorithm grouped the selected stocks into three distinct clusters based on the correlation of their log returns. Table~\ref{tab:cluster_allocations} presents the stock allocations for each cluster.

\begin{table}[h!]
\centering
\begin{tabular}{ll}
\hline
\textbf{Cluster ID} & \textbf{Stocks} \\
\hline
1 & \texttt{AMZN}, \texttt{V} \\
2 & \texttt{AAPL}, \texttt{NVDA}, \texttt{META}, \texttt{UNH} \\
3 & \texttt{MSFT}, \texttt{GOOGL}, \texttt{TSLA}, \texttt{JPM} \\
\hline
\end{tabular}
\caption{Stock clusters derived from K-Means clustering.}
\label{tab:cluster_allocations}
\end{table}

\subsection{Correlation Matrix Heatmap}

\begin{figure}[h!]
\centering
\includegraphics[width=0.8\textwidth]{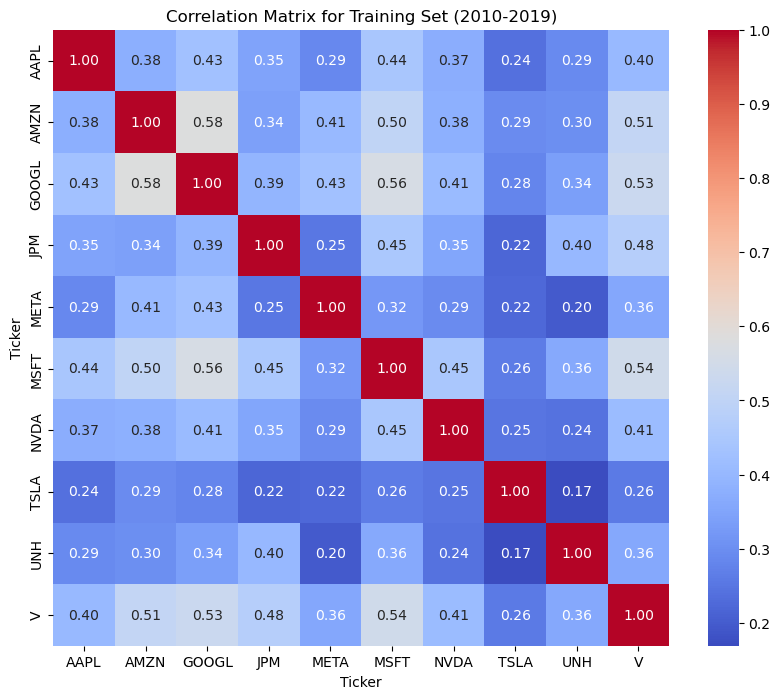}
\caption{Correlation matrix heatmap for selected stocks.}
\label{fig:correlation_heatmap}
\end{figure}

Figure~\ref{fig:correlation_heatmap} visualizes the correlation matrix of log returns, highlighting the relationships between the selected stocks.

\subsection{Portfolio Optimization Results}

The optimized weights for Cluster 2, which contains \texttt{AAPL}, \texttt{NVDA}, \texttt{META}, and \texttt{UNH}, are presented in Table~\ref{tab:optimized_weights}. The weights were calculated using Sharpe ratio optimization.

\begin{table}[h!]
\centering
\begin{tabular}{lc}
\hline
\textbf{Stock} & \textbf{Weight} \\
\hline
\texttt{AAPL}  & 0.0570 \\
\texttt{NVDA}  & 0.7118 \\
\texttt{META}  & 0.0000 \\
\texttt{UNH}   & 0.2311 \\
\hline
\end{tabular}
\caption{Optimized weights for stocks in Cluster 2.}
\label{tab:optimized_weights}
\end{table}

\subsection{Backtesting Performance}

\begin{figure}[h!]
\centering
\includegraphics[width=0.8\textwidth]{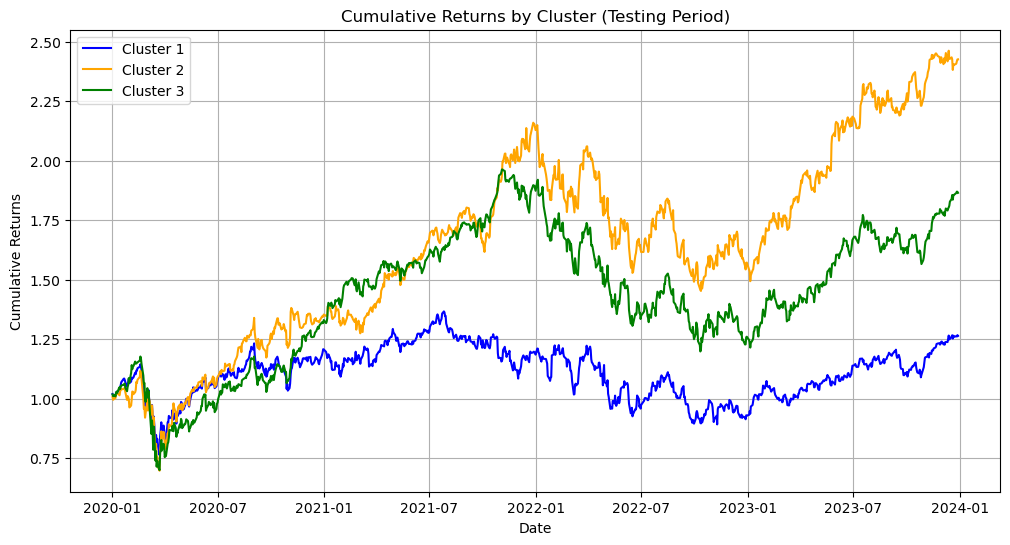}
\caption{Cumulative Returns: Clusters vs. Benchmark Portfolio.}
\label{fig:clusters_vs_benchmark}
\end{figure}

\begin{figure}[h!]
\centering
\includegraphics[width=0.8\textwidth]{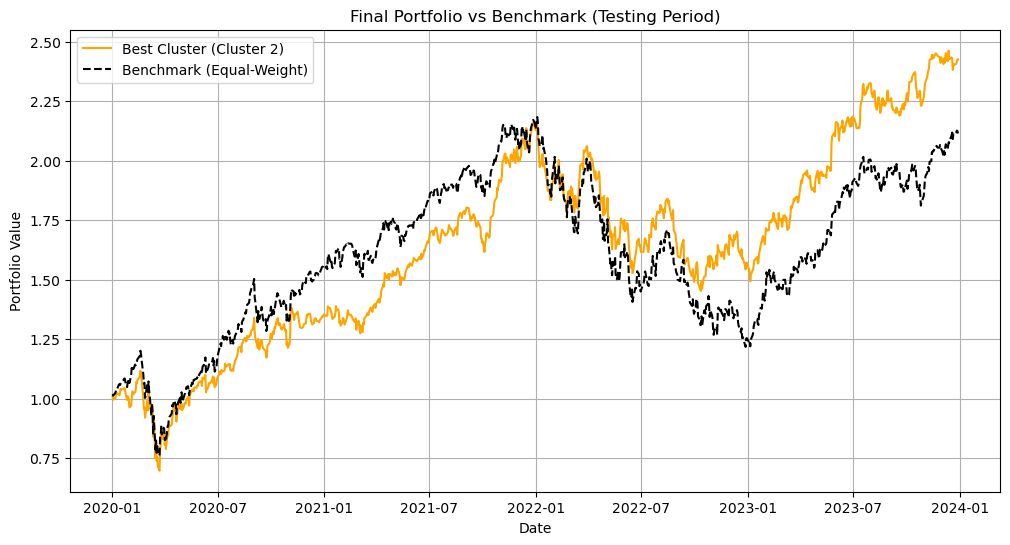}
\caption{Cumulative Returns: Best Cluster (Cluster 2) vs. Benchmark Portfolio.}
\label{fig:best_cluster_vs_benchmark}
\end{figure}

The cumulative performance of the cluster portfolios (Clusters 1, 2, and 3) is compared with the benchmark portfolio in Figure~\ref{fig:clusters_vs_benchmark}. Cluster 2, which demonstrated superior performance, is further analyzed against the benchmark portfolio in Figure~\ref{fig:best_cluster_vs_benchmark}.

\subsection{Performance Metrics}

The performance metrics for the selected cluster portfolio (Cluster 2) and the benchmark portfolio are summarized in Table~\ref{tab:performance_metrics}. The selected cluster portfolio significantly outperformed the benchmark portfolio in terms of total return, annualized return, and Sharpe ratio.

\begin{table}[h!]
\centering
\begin{tabular}{lcccc}
\hline
\textbf{Portfolio} & \textbf{Total Return (\%)} & \textbf{Annualized Return (\%)} & \textbf{Volatility (\%)} & \textbf{Sharpe Ratio} \\
\hline
Selected Cluster & 140.98 & 24.67 & 30.60 & 0.84 \\
Benchmark        & 107.59 & 20.09 & 30.09 & 0.73 \\
\hline
\end{tabular}
\caption{Performance metrics of the selected cluster portfolio and benchmark portfolio.}
\label{tab:performance_metrics}
\end{table}

\textbf{Interpretation:}
\begin{itemize}
    \item The \textbf{Selected Cluster Portfolio (Cluster 2)} achieved a total return of 140.98\%, outperforming the benchmark portfolio's total return of 107.59\%.
    \item The annualized return for the selected cluster portfolio was 24.67\%, higher than the benchmark portfolio's annualized return of 20.09\%.
    \item The selected cluster portfolio exhibited slightly higher volatility (30.60\%) compared to the benchmark portfolio (30.09\%), reflecting similar risk exposure.
    \item The Sharpe ratio of the selected cluster portfolio (0.84) exceeded that of the benchmark portfolio (0.73), indicating superior risk-adjusted performance.
\end{itemize}

These findings suggest that the clustering-based approach, combined with Sharpe ratio optimization, can substantially enhance portfolio performance while maintaining a reasonable risk level.
\section{Conclusion and Future Work}

This study demonstrates the efficacy of a clustering-based approach combined with Sharpe ratio optimization for portfolio construction. The integration of K-Means clustering and portfolio optimization enabled the identification of high-performing asset clusters and allocation of weights that maximize risk-adjusted returns. Through backtesting on a dataset spanning 2010 to 2024, the results show that:

\begin{itemize}
    \item Cluster 2, consisting of \texttt{AAPL}, \texttt{NVDA}, \texttt{META}, and \texttt{UNH}, was identified as the best-performing cluster during the testing period (2020–2024).
    \item The optimized weights for Cluster 2 were heavily skewed towards \texttt{NVDA} (71.18\%) and \texttt{UNH} (23.11\%), with minimal allocation to \texttt{AAPL} (5.70\%) and no allocation to \texttt{META}.
    \item The selected portfolio achieved a total return of 140.98\%, an annualized return of 24.67\%, and a Sharpe ratio of 0.84, outperforming the benchmark portfolio with a total return of 107.59\%, an annualized return of 20.09\%, and a Sharpe ratio of 0.73.
\end{itemize}

These findings highlight the potential of combining clustering techniques with portfolio optimization to achieve superior performance in dynamic market conditions.

\subsection*{Future Work}
While this study offers promising results, several areas remain for further exploration:

\begin{itemize}
    \item \textbf{Incorporating Advanced Machine Learning Models:} Future research could explore the use of advanced machine learning models, such as neural networks or ensemble methods, for more accurate return predictions.
    \item \textbf{Dynamic Rebalancing:} Investigating the effects of dynamic rebalancing strategies based on periodic updates to clustering and optimization.
    \item \textbf{Sector-Based Analysis:} Expanding the dataset to include a broader range of sectors to evaluate the generalizability of the proposed framework.
    \item \textbf{Incorporating Macroeconomic Indicators:} Integrating macroeconomic variables, such as interest rates or GDP growth, into the clustering and optimization process to enhance predictive accuracy.
    \item \textbf{Robustness Testing:} Conducting robustness checks with alternative clustering algorithms (e.g., hierarchical clustering) and optimization objectives (e.g., minimizing Conditional Value at Risk).
\end{itemize}

By addressing these areas, future studies can further refine the methodology and extend its applicability to real-world financial markets.

\section*{CRediT authorship contribution statement}
\textbf{Keon Vin Park:} Conceptualization, Methodology, Software, Writing – original draft, Writing – review \& editing.

\section*{Declaration of Competing Interest}
The authors declare that they have no known competing financial interests or personal relationships that could have appeared to influence the work reported in this paper.

\section*{Acknowledgements}
This work was partly supported by Institute of Information \& communications Technology Planning \& Evaluation (IITP) grant funded by the Korea government(MSIT) [NO.RS-2021-II211343, Artificial Intelligence Graduate School Program (Seoul National University)].

\section*{Data availability}
The dataset used in this study, consisting of daily closing prices for the S\&P 500 index, is available for download on Kaggle. Interested researchers can access the dataset through the following link: \url{https://www.kaggle.com/datasets/andrewmvd/sp-500-stocks}. This resource provides a comprehensive historical record of stock prices required to replicate the analyses presented in this paper.

\bibliography{sample}

\end{document}